\pdfoutput=1

\documentclass[11pt]{article}

\usepackage[]{emnlp2021}

\usepackage{times}
\usepackage{latexsym}
\usepackage{pythonhighlight}
\usepackage{setspace}
\usepackage{graphicx}
\usepackage{hyperref}
\usepackage{bm}
\usepackage{xcolor}
\usepackage{adjustbox}
\usepackage{diagbox}
\usepackage[english]{babel}
\usepackage{pdfpages} 
\usepackage{subfigure}
\usepackage{amsmath}
\usepackage{amssymb}
\newcommand{\Po}{SensePOLAR} 
\newcommand{\Poo}{SensePOLAR } 

\usepackage[T1]{fontenc}

\usepackage[utf8]{inputenc}

\usepackage{microtype}

\title{\Po: Word sense aware interpretability for pre-trained contextual word embeddings}

\author{Jan Engler\thanks{~~Equal contribution} \\
  RWTH Aachen \\
  \texttt{jan.engler@rwth-aachen.de} \\\And
  Sandipan Sikdar\footnotemark[1] \\
  L3S Research Center \\ 
  \texttt{sandipan.sikdar@l3s.de} \\ \AND
  \newline
  Marlene Lutz \\
  University of Mannheim \\
  \And 
  Markus Strohmaier \\
  University of Mannheim, GESIS, CSH Vienna \\
  \hspace{-55mm}\texttt{\{marlene.lutz, markus.strohmaier\}@uni-mannheim.de}\\
  }

\begin{document}
\maketitle
\begin{abstract}
Adding interpretability to word embeddings represents an area of active research in text representation. Recent work has explored the potential of embedding words via so-called \emph{polar} dimensions (e.g. good vs. bad, correct vs. wrong). Examples of such recent approaches include SemAxis, POLAR, FrameAxis, and BiImp. Although these approaches provide interpretable dimensions for words, they have not been designed to deal with polysemy, i.e. they can not easily distinguish between different senses of words.
To address this limitation, we present \Po, an extension of the original POLAR framework that enables word-sense aware interpretability for pre-trained \emph{contextual} word embeddings.
The resulting \emph{interpretable} word embeddings achieve a level of performance that is comparable to original contextual word embeddings across a variety of natural language processing tasks including the GLUE and SQuAD benchmarks. Our work removes a fundamental limitation of existing approaches by offering users sense aware interpretations for contextual word embeddings.  

\end{abstract}

\begin{figure*}[t]
	\centering
	\includegraphics[scale=0.425]{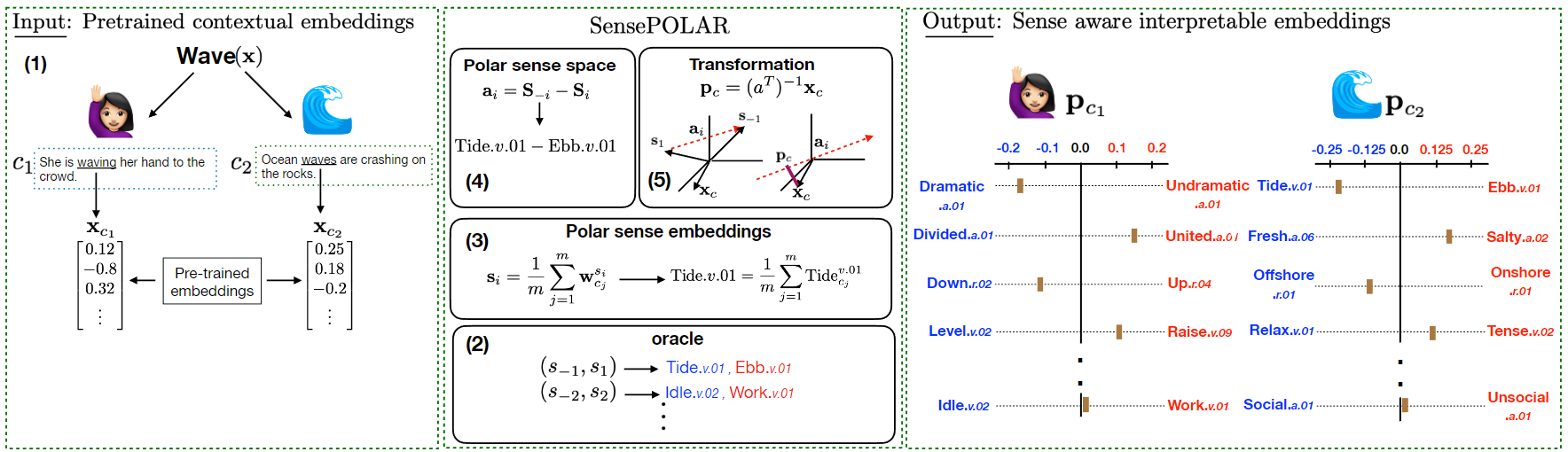}
	\caption{\Poo overview. Pre-trained contextual word embeddings are transformed into an interpretable space where the word's semantics are rated on scales individually encoded by opposite senses such as ``good''$\leftrightarrow$``bad''. The scores across the dimensions are representative of the strength of relationship (between word and dimension) which allows us to rank the dimensions and thereby identify the most discriminative dimensions for a word.
	In this example, the word ``wave'' is used in two senses: \emph{hand waving} and \emph{ocean wave}. \Poo not only generates dimensions that are representative of individual contextual meanings, the alignment to the respective sense spaces also aligns well with human judgement. \Poo generates neutral scores for dimensions not related to the word in the given context (e.g., ``idle''$\leftrightarrow$``work'', ``social''$\leftrightarrow$``unsocial''). We follow the WordNet convention to represent a particular sense of a word. For example, ``Tide.v.01'' represents the word ``tide'' in the sense of \emph{surge (rise or move forward)}. 
	}
	\vspace{-3mm}
	\label{fig:overview}
\end{figure*}

\section{Introduction}

The overwhelming success of deep neural networks (DNN) in the last decade has been accompanied by increasing concerns about the lack of  \textit{interpretability}~\cite{ribeiro2016why}.
This problem is amplified in the area of Natural Language Processing (NLP) where \textit{word embeddings} are used as input to machine learning models instead of more classical, understandable features.
Traditional (\textit{static}) word embedding models like Word2Vec \cite{mikolov2013efficient} or Glove \cite{pennington2014glove}, that create one embedding for each word, are currently being replaced by \textit{contextual} word embedding models like 
BERT~\cite{devlin2018bert} 
which have achieved competitive performance in NLP benchmarks such as GLUE~\cite{wang2018glue} and SQuAD~\cite{rajpurkar2016squad}. 

To improve interpretability, recent approaches such as SemAxis \cite{an2018semaxis}, POLAR \cite{mathew2020polar}, FrameAxis \cite{kwak2021frameaxis}, and
BiImp \cite{csenel2022learning} have explored the potential of embedding words via polar dimensions (e.g. good vs. bad, correct vs. wrong). While these approaches have been useful for interpreting word vectors, they have not been designed to deal with polysemy, i.e. multiple senses of words.

\noindent\textbf{Objective:}
Addressing polysemy, in this paper we aim to enable \emph{word-sense aware} interpretability for pre-trained \textit{contextual} word embeddings. 

\noindent\textbf{Approach:}
We base our approach on the original POLAR framework \cite{mathew2020polar} and  
the idea of semantic differentials~\cite{osgood1957measurement}, which are psychometric scales between two antonym words, e.g. ``right'' $\leftrightarrow$ ``wrong''. 
\Poo extends POLAR~\cite{mathew2020polar} to contextual word embeddings, and defines polar \textit{sense} instead of polar \textit{word} scales.
This enables \Poo to offer polar dimensions that distinguish between the \textit{correctness} sense of ``right'' and the \textit{direction} sense of ``right'', for example. 



\noindent\textbf{Results:}
SensePOLAR enables word sense aware \emph{interpretability} of contextual embeddings by selecting polar sense dimensions that align reasonably well with human judgements, as demonstrated in survey experiments.
SensePOLAR exhibits competitive \emph{performance} on various NLP tasks where it is used as input features for a separate model (feature-based approach) as well as directly integrated in the model itself (fine-tuning approach).


\noindent\textbf{Contributions:} \Poo introduces the notion of \emph{sense aware} interpretations. To the best of our knowledge, \Poo represents the first (semi-) supervised method that enables word sense aware interpretability for contextual word embeddings. \Poo is publicly available\footnote{\url{https://github.com/JanEnglerRWTH/SensePOLAR}}. 

\section{\Poo}
The key idea of \Poo is to transform pre-trained word embeddings into an interpretable, \emph{sense aware} space.
In this space, each dimension represents a scale on which words are rated, inspired by the semantic differential technique~\cite{osgood1957measurement}.
In a departure from the existing approaches, we define opposite \textit{senses} for the poles of these scales (e.g. ~{``left direction'' $\leftrightarrow$ ``right direction''}), as opposed to opposite words (e.g. ~{``left'' $\leftrightarrow$ ``right''}), as used in ~\citet{mathew2020polar}.

Given a contextual word embedding model $\mathcal{M}$, the interpretable embeddings are obtained through the following steps. 1) We use $\mathcal{M}$ to obtain the (non-interpretable) contextual embedding space. 2) We obtain polar senses with contextual information from an oracle. 3) We proceed with generating representative sense embeddings from which we 4) construct the interpretable polar sense space. 5) The original embedding is transformed into the polar sense space, which enables interpretation with regard to opposite sense pairs. 
We illustrate each step in figure~\ref{fig:overview} and elaborate them next. 

\noindent\textbf{1. Obtaining contextual embeddings:} 
To obtain the embedding of a particular word, we forward the word with its context, i.e. an example sentence, to the embedding model $\mathcal{M}$. The embedding of the corresponding word can then be retrieved from the output of $\mathcal{M}$. Because most models deploy subword tokenization algorithms, such as WordPiece ~\cite{wu2016google}, embeddings of only \emph{subword} tokens, rather than entire words, are generated by the contextual embedding models. This provides for obtaining representations for out-of-vocabulary words but, at the same time, makes embeddings of even common words not directly available. Following existing literature~\cite{mccormick2019tutorial,bommasani-etal-2020-interpreting}, we compute the embedding of a word by averaging over the embeddings of the constituent tokens. 

\noindent\textbf{2. Selecting opposite polar senses:} 
Each dimension in the interpretable space corresponds to a scale spanned by opposite polar senses, which we define as a \emph{polar sense dimension}. 
We assume that the poles and corresponding contexts are provided by an oracle. In this paper, we use WordNet~\cite{miller1995wordnet} as an oracle, since the database already provides senses, contexts and antonyms for many words. Each sense of a word is represented by a unique identifier, e.g. ``Right.r.0'' (a convention followed in WordNet) encodes ``right'' in the sense of \emph{direction}.
From over 6000 sense-antonym pairs that are available in WordNet, we use only a subset (1763) that are annotated with example sentences for both words. After various post-processing steps (cf. Appendix), these example sentences are used as context in step 3.

\noindent\textbf{3. Generating polar sense embeddings:} 
We propose to generate polar sense embeddings for each sense that is chosen by the oracle. Let $w$ denote the word of interest and $s$ the word-sense. Furthermore, let $C_s = \{c_1, ..., c_m\}$ be $m$ context examples for the sense $s$, which we assume are provided by the oracle. In each context $c \in C_s$, the word $w$ is used in the sense $s$, e.g. ``A strange sound came from the right side.'' for the word ``right'' in the sense of \textit{direction} (i.e., we intend to embed ``Right.r.04''). We create polar sense embeddings in two steps. First, we input $m$ context examples for a sense $s$ to the embedding model $\mathcal{M}$ and retrieve an embedding $\mathbf{w}_{c}^s \in \mathbb{R}^d$ of the word $w$ for each context $c \in C_s$. If the word $w$ consists of several subword tokens, the individual subword embeddings are averaged. We also allow for senses consisting of multiple words, e.g. ``keep track'' $\leftrightarrow$ ``lose track'', where we again average the embeddings of the individual tokens. Second, we compute the average of the contextual word embeddings per sense and define it as the sense embedding $\mathbf{s} \in \mathbb{R}^d$:
\begin{equation} \label{eq:1}
	\mathbf{s} = \frac{1}{m} \sum_{j=1}^m \mathbf{w}_{c_j}^s
\end{equation}

This is a rather straightforward way to represent individual senses of words in a (semi-) supervised manner.
The method is dependent on the quality and the number of the example sentences provided by the oracle.
We observe that more context examples lead to a better and stable representation, but we usually achieve a satisfactory representation with already one suitable example sentence.
This is motivated by the observations in~\citet{reif2019visualizing} which provide strong evidence that BERT positions the embeddings of senses in individual clusters in space and that these clusters are usually sufficiently spatially separated from each other. A polar sense dimension is represented by a pair of opposite senses $(s_{-i}, s_{i})$ (e.g., ``Right.a.02'', ``Wrong.a.01'').

\noindent\textbf{4. Constructing a polar sense space:} 

\noindent Given $n$ polar sense dimensions ${\mathbf{S}=((s_{-1},s_{1}), ..., (s_{-n},s_{n}))}$ with their contexts $\mathbf{C}=((C_{s_{-1}}, C_{s_{1}}), (C_{s_{-n}}, C_{s_{n}}))$, we compute  the polar sense embedding $\mathbf{s}_i$ for each sense $s_i$ and corresponding context $C_{s_i}$, following equation~\ref{eq:1}.

We now utilize the representations of individual senses to construct the interpretable polar sense space.
Each polar sense dimension $(s_{i},s_{-i}) \in \mathbf{S}$ defines an interpretable scale, which is encoded by the direction vector $\mathbf{a_i}$, defined as follows:

\begin{equation}
	\mathbf{a}_i = \mathbf{s}_{-i} - \mathbf{s}_{i}
\end{equation}


The direction vectors for all polar sense dimensions are then stacked to obtain the change of basis matrix $\mathbf{a} \in \mathbb{R}^{n \times d}$ for the interpretable polar sense space. 

\noindent\textbf{5. Transformation to interpretable embeddings:} Finally, an embedding of a word $x$ in a context $c$,  $\mathbf{x}_{c}$ can be transformed into the polar sense space in the following way. Given $\mathbf{a}$ represents the change of basis matrix, we can compute the polar sense embedding $\mathbf{p}_{c}$ following the rules of linear algebra:

\begin{align}
    \mathbf{a}^T \;\mathbf{p}_{c} =& \;  \mathbf{x}_{c} \\
	\mathbf{p}_{c} =& \; (\mathbf{a}^T)^{-1} \;\;  \mathbf{x}_{c}
\end{align}


The inverse of $\mathbf{a}^T$ is computed by the Moore-Penrose generalized inverse~\cite{ben2003generalized}.
The resulting contextual word embedding $\mathbf{p}_{c}$ in the polar sense space is of dimension $n \times 1$.
The absolute value across axis $\mathbf{a_i}$ corresponds to the word's rating on the scale between the polar senses $(s_{-i},s_{i})$ and the sign represents the direction of alignment to a particular pole.
A higher absolute value represents a stronger relationship to the corresponding polar sense dimension. This allows us to obtain the most expressive polar sense dimensions for a given word and context.

\noindent\textbf{Normalization:} As a post-processing step, we average the word embeddings of all words (from a corpus) to get the average-word embedding in our interpretable space and subtract this average word embedding from each embedding when analyzing interpretability. This also allows us to deal with the \emph{anisotropic} nature of contextual word embeddings~\cite{ethayarajh-2019-contextual} whereby the embeddings are not randomly distributed but rather lay on a high-dimensional cone in space.

\section{Evaluation}

Note that while \Poo allows for deployment across any contextual word embedding model, in this work, we consider  BERT~\cite{devlin2018bert} as our model for illustration. We consider a BERT-base model which utilizes 12 transformer~\cite{vaswani2017attention} encoder layers and generates embeddings of size 768. The pre-trained BERT-base model was downloaded from Huggingface\footnote{\url{https://huggingface.co/docs/transformers/model_doc/bert}, we used ``bert-base-uncased''}.
In addition, we use WordNet as our oracle with 1763 polar sense pairs. 

\subsection{Performance on downstream tasks}
The goal of \Poo is to add interpretability to word embeddings without major losses in performance.
Hence, we evaluate \Poo on a wide range of NLP downstream tasks. We investigate whether replacing the original BERT embeddings with \Poo embeddings has any effect on performance.

\begin{table}[]
\begin{adjustbox}{width=\columnwidth,center}
\begin{tabular}{c|cc|cc}
\hline
ML Model & \multicolumn{2}{c|}{SVM}               & \multicolumn{2}{c}{FFN}               \\ \hline
Task     & \multicolumn{1}{c|}{Base}  & \Po & \multicolumn{1}{c|}{Base}  & \Po \\ \hline
Sport    & \multicolumn{1}{c|}{0.941} & 0.935$\textcolor{red}{\downarrow 0.6\%}$     & \multicolumn{1}{c|}{0.961} & 0.956$\textcolor{red}{\downarrow 0.5\%}$     \\ 
Religion & \multicolumn{1}{c|}{0.891} & 0.848$\textcolor{blue}{\uparrow 4.8\%}$     & \multicolumn{1}{c|}{0.880} & 0.894$\textcolor{blue}{\uparrow 1.6\%}$     \\ 
Computer & \multicolumn{1}{c|}{0.770} & 0.750$\textcolor{red}{\downarrow 2.6\%}$     & \multicolumn{1}{c|}{0.763} & 0.727$\textcolor{red}{\downarrow 4.7\%}$     \\ \hline
\end{tabular}
\end{adjustbox}
\caption{\label{tab:feature} Performance of the original BERT (Base) embeddings and \Poo  embeddings on feature-based tasks with a support vector machine (SVM) and a feed-forward neural network (FFN) classifier. \Poo achieves performance comparable to the original BERT embeddings across all three tasks.}
\vspace{-4mm}
\end{table}

\subsubsection{Feature-based tasks}
We analyze the effectiveness of \Poo embeddings in a  ``classical'' NLP pipeline, where word embeddings are generated beforehand and are used as input-features to a \textit{separate} machine learning model. We consider a binary text classification task utilizing the 20 Newsgroups dataset~\cite{lang1995newsweeder}. The dataset consists of $\sim 20K$ news articles covering 20 types of news. Our experiment follows the structure of \citet{panigrahi2019word2sense}, where we only consider the topics sports, computer and religion.
For each topic, an article must be classified into one of two categories (``baseball'' or ``hockey'' for sports, ``IBM'' or ``Apple'' for computer, ``christianity'' or ``atheism'' for religion).
In table~\ref{tab:feature}, we present the results in terms of accuracy across the three tasks. We use a support vector machine (SVM) and a 2-layer feed-forward neural network (FFN) as classifier models, which use the BERT and \Poo embeddings as features. Across all three tasks, \Poo achieves a level of performance that is comparable to the original embeddings.

\subsubsection{Fine-tuning tasks}

\noindent\textbf{Integrating \Poo into fine-tuned models:} The models achieving state-of-the-art performances on different NLP tasks usually deploy a task specific network layer (usually a feed-forward network) on top of the embedding layers. The embedding layers and the task specific layers are then fine-tuned on the task specific dataset. 
Consequently, \Poo embeddings need to be computed considering the fine-tuned version of the embeddings rather than the original pre-trained version. 
In this particular setting, we propose to utilize the embedding layer of the fine-tuned model (instead of the original pre-trained version) to construct the polar sense space.
Given an input text, each token (including the \verb*|[CLS]| token) can then be transformed to a corresponding \Poo embedding. Because of the dimensionality mismatch between the original embedding and the transformed \Poo embedding, we replace the first layer of the task specific feed-forward network 
and re-fine-tune it on the task specific dataset. Note that the weights of the underlying embedding model are frozen during this re-fine-tuning procedure. This is computationally inexpensive as only the task specific layers need to be trained, which are often just 1 or 2-layered feed-forward network. 

\noindent\textbf{Question answering:} This task deals with locating an answer to a question in a given paragraph and is often referred to as a reading comprehension task. We consider the SQuAD benchmark, including both SQuAD1.1~\cite{rajpurkar2016squad} and SQuAD2.0~\cite{rajpurkar2018know} versions. The BERT-based QA model consists of the embedding module followed by a span-classification head, which is a 1-layer feed-forward network. The model takes both the question text and the passage text as input. The \verb*|[CLS]| token (a special token generated by BERT for classification tasks) obtained from the embedding module is then passed onto the span-classification head, which predicts the start and the end position of the span in the text passage that contains the answer. 

\begin{table}[]
\begin{adjustbox}{width=\columnwidth,center}
\begin{tabular}{c|cc|cc}
\hline
       & \multicolumn{2}{c|}{SQuAD 1.1}         & \multicolumn{2}{c}{SQuAD 2.0}         \\ \hline
Metric & \multicolumn{1}{c|}{Base}  & \Po & \multicolumn{1}{c|}{Base}  & \Po \\ \hline
EM     & \multicolumn{1}{c|}{86.92} & 86.85$\textcolor{red}{\downarrow 0.07\%}$     & \multicolumn{1}{c|}{80.88} & 81.06$\textcolor{blue}{\uparrow 0.22\%}$     \\ 
F1     & \multicolumn{1}{c|}{93.15} & 93.12$\textcolor{red}{\downarrow 0.03\%}$     & \multicolumn{1}{c|}{83.87} & 83.89$\textcolor{blue}{\uparrow 0.02\%}$     \\ \hline
\end{tabular}
\end{adjustbox}
\caption{Results of fine-tuned BERT embeddings and with \Poo transformed embeddings on the SQuAD benchmark. 
The results are competitive and even improve marginally after applying \Po.}
\vspace{-4mm}
\label{table:squad}
\end{table}

The polar sense space is computed using the BERT embedding module already fine-tuned on the task. The \verb*|[CLS]| token is then transformed into the interpretable space before being passed on to the span-classification head. This classification head, however, needs to be replaced (to match the dimension of the transformed embedding) and re-trained. 
In table~\ref{table:squad}, we report the exact match (EM) and F1 scores with the original BERT (base) and the \Poo model. \Poo again achieves comparable performance, even marginally outperforming the base model for SQuAD2.0. 

\begin{figure*}[t]
	\begin{center}
	    \subfigure[Benefits of sense dimensions]{\includegraphics[scale=0.22]{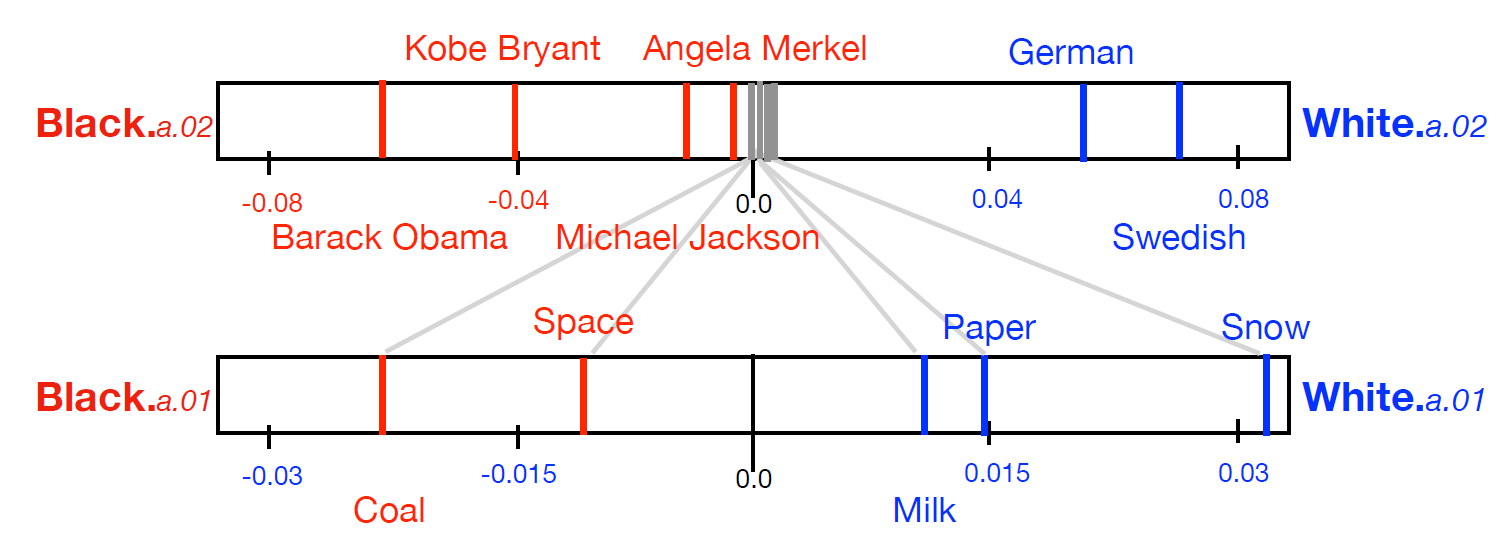}}
    \subfigure[Connotative meanings across dimensions]{\includegraphics[scale=0.22]{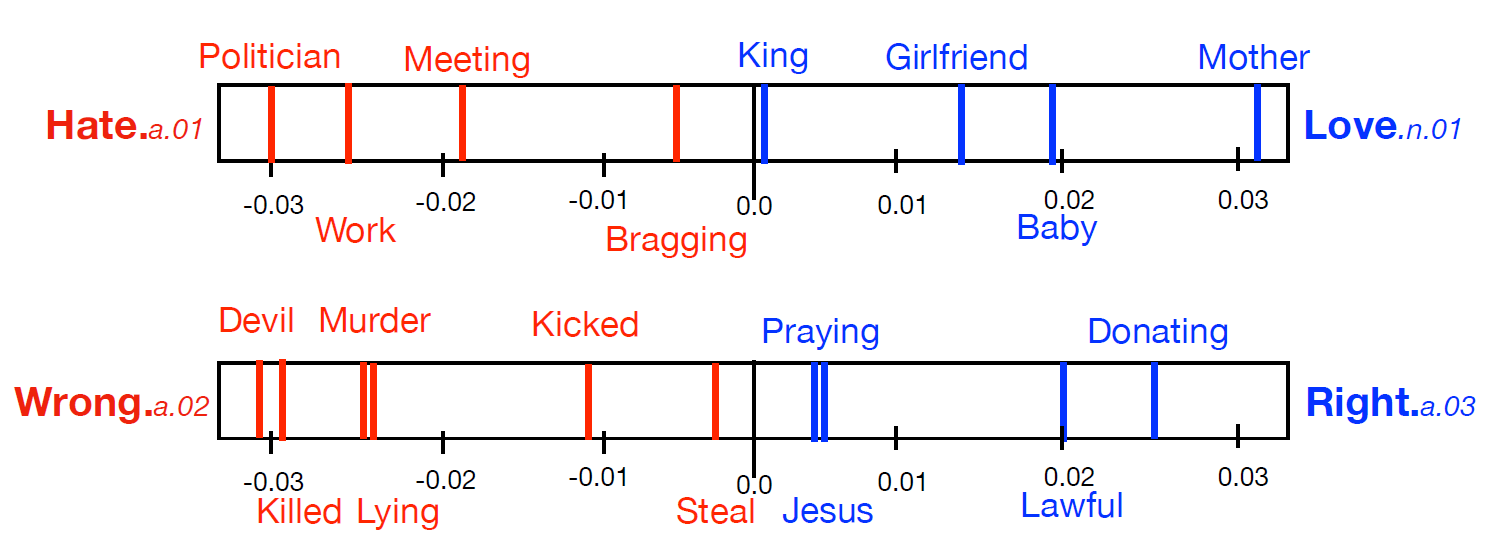}}
	\end{center}
	\vspace{-4mm}
	\caption{Illustration of polar sense dimensions. (a) \Poo allows for interpretability along multiple senses. ``black''$\leftrightarrow$``white'' in the sense of \emph{ethnicity} (top) can be differentiated from ``black''$\leftrightarrow$``white'' in the sense of \emph{color} (bottom). Words like ``snow''or ``coal'' - which are not semantically related to ethnicity - score neutral on the upper scale while being clearly distinguishable on the lower scale. (b) The connotative meanings of words can also be investigated through \Po. For example, ``politician'' is associated with ``hate'' while ``mother'' is associated with ``love''.}
	\vspace{-4mm}
	\label{fig:coldhot}
\end{figure*}

\noindent\textbf{Natural language understanding:} We utilize the General Language Understanding Evaluation (GLUE) benchmark, which is designed for comparing models on the task of natural language understanding (NLU). 
It consists of nine tasks that cover a diverse range of text genres, dataset sizes, and degrees of difficulty~\cite{wang2018glue}. We point the reader to the original paper by~\citet{wang2018glue} for a general overview of the tasks. To evaluate \Po, we follow a similar procedure to the previous question answering task. 
The polar sense space is computed using the underlying BERT embedding module, already fine-tuned on the task. 
This is followed by transforming the \verb*|[CLS]| token into the interpretable polar sense space. The feed-forward layers on top are then replaced and re-trained. 
\begin{table}[h!]
\begin{adjustbox}{width=\columnwidth,center}
	\centering
	\begin{tabular}{c| c | c | c |c } 
		\hline
		Task &Train size& Metric& Base&\Po \\
		\hline
		CoLa& 8.5k & Matthew's corr. & 56.62 & 55.05 $\textcolor{red}{\downarrow 2.77\%}$\\
		\hline
		SST-2& 67k & Accuracy & 91.51 & 91.40 $\textcolor{red}{\downarrow 0.12\%}$\\
		\hline
		MRPC& 3.7k & Accuracy  & 84.31 & 82.84 $\textcolor{red}{\downarrow 1.74\%}$\\
		& &  F1 & 89.00 & 87.41 $\textcolor{red}{\downarrow 1.79\%}$\\
		\hline
		STS-B& 7k & Person corr. & 89.03 & 84.17 $\textcolor{red}{\downarrow 5.46\%}$\\
		\hline
		QQP& 364k & Accuracy  & 90.59& 90.15 $\textcolor{red}{\downarrow 0.49\%}$ \\
		&  &  F1 & 87.29 & 86.82 $\textcolor{red}{\downarrow 0.54\%}$\\
		\hline
		MNLI& 393k &Accuracy  & 84.49 & 84.04 $\textcolor{red}{\downarrow 0.53\%}$\\
		\hline
		QNLI& 105k & Accuracy & 91.54 &91.58 $\textcolor{blue}{\uparrow 0.04\%}$ \\
		\hline
		RTE& 2.5k & Accuracy & 63.18 & 59.93 $\textcolor{red}{\downarrow 5.14\%}$\\
		\hline
		WNLI& 634 & Accuracy & 56.34 & 56.34 $\textcolor{blue}{\uparrow}\textcolor{red}{\downarrow} 0\%$\\
		\hline
	\end{tabular}
	\end{adjustbox}
	\caption{Comparison of the fine-tuned BERT model and the re-fine-tuned BERT model with \Poo embeddings. Mostly, comparable performance is achieved. Slightly worse performance is achieved for tasks with smaller training datasets.}
	\label{table:glue}
	\vspace{-3mm}
\end{table}
In table~\ref{table:glue}, we report the results on GLUE tasks with both the original BERT (Base) and the \Poo embeddings. \Poo achieves competitive performances across all the tasks. 

The results indicate that \Poo is able to achieve interpretability without compromising performance on downstream tasks.

\subsection{Interpretability}
We turn our attention to evaluating the interpretability of \Po. 

\begin{figure*}[t]
	\begin{center}
	    \subfigure[Run]{\includegraphics[scale=0.17]{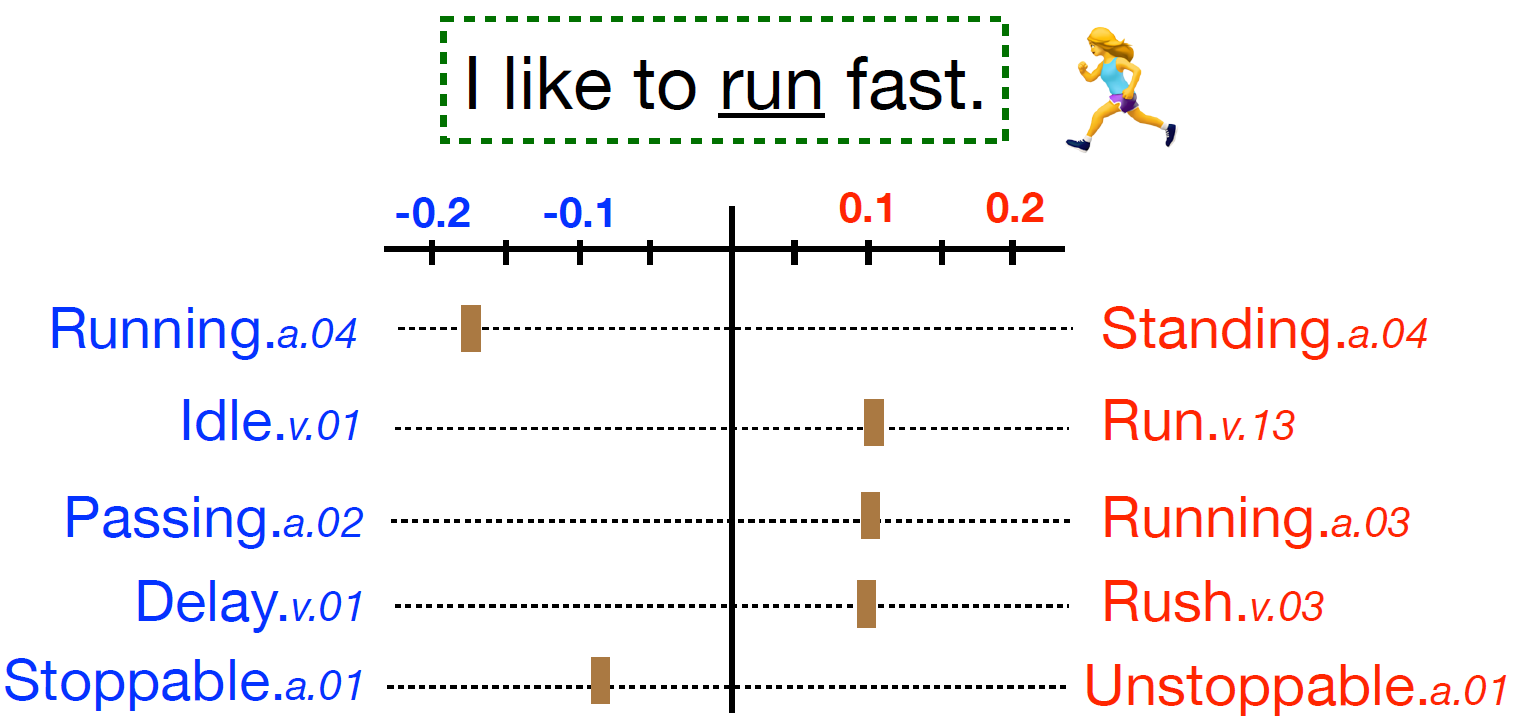}}
    \subfigure[Music]{\includegraphics[scale=0.18]{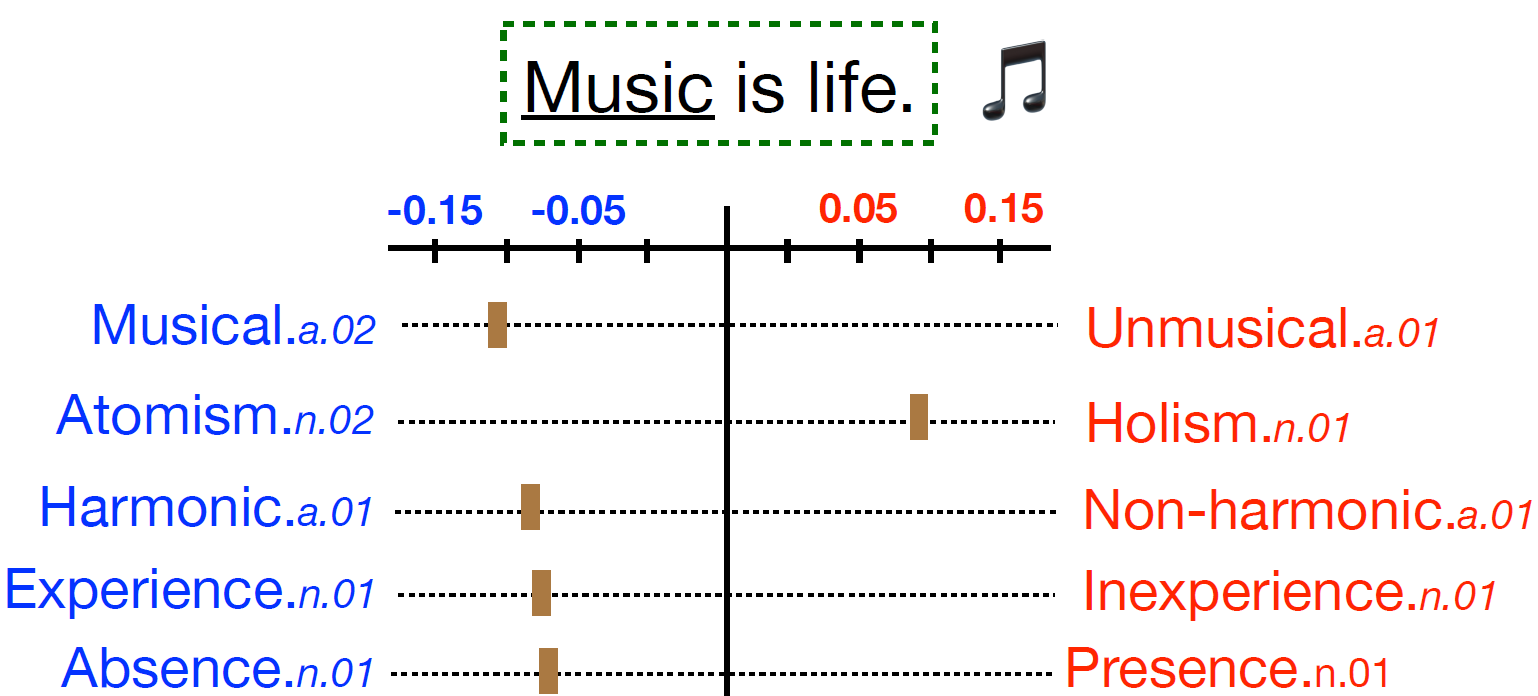}}
    \subfigure[Right]{\includegraphics[scale=0.16]{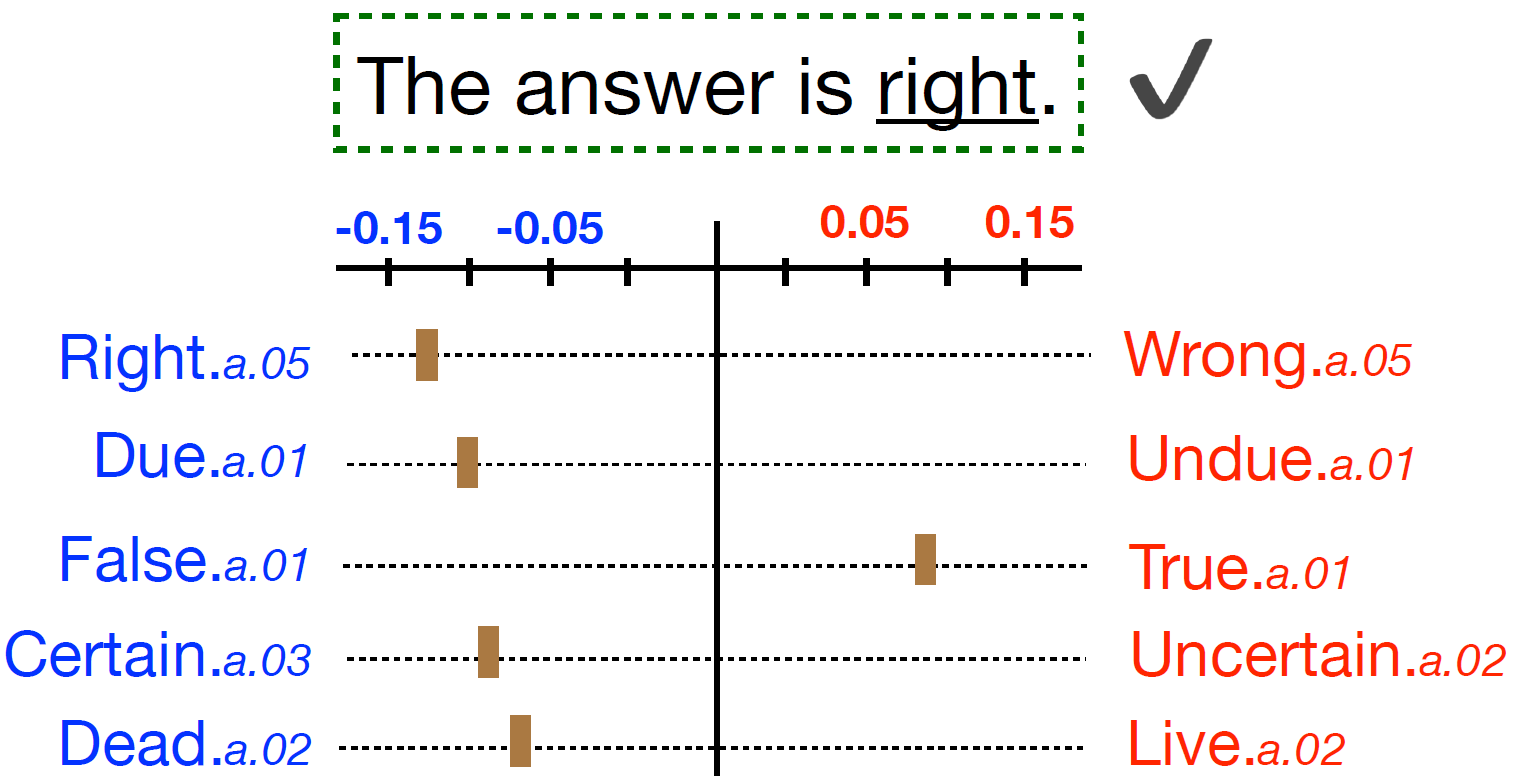}}
    \subfigure[Happy]{\includegraphics[scale=0.18]{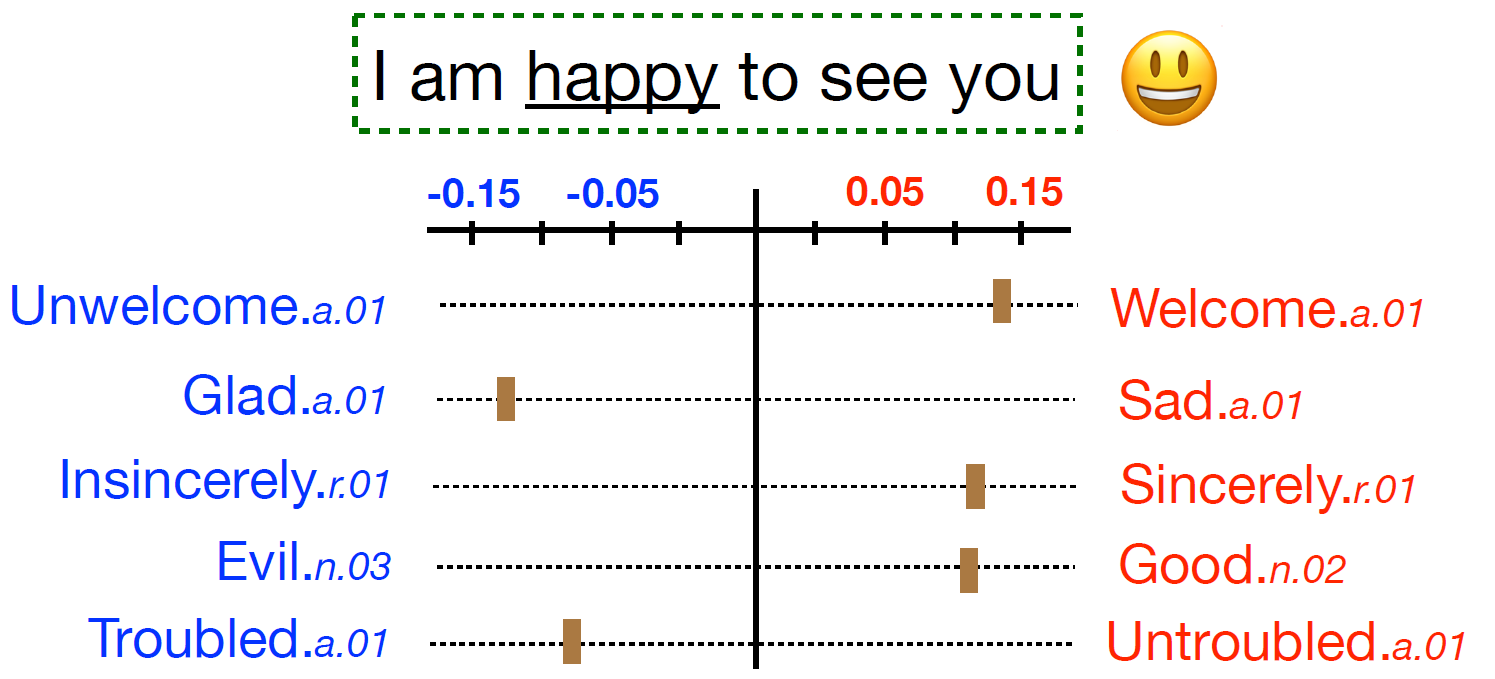}}
	\end{center}
	\vspace{-5mm}
	\caption{\label{fig:polar_c_examples} Illustration of \Poo embeddings. We show the top 5 dimensions as selected by \Poo for exemplary words. The pre-trained embeddings are obtained using BERT. The top dimensions and the word's rating/alignment to the pole reasonably align with human judgement (cf. 	table \ref{table:conditional}).}
	\vspace{-4mm}
\end{figure*}

\noindent\textbf{Qualitative analysis:} 
We transform the embeddings of the words into a polar sense space and analyze the position/rating (determined by the signed value on that dimension) of different words on selected dimensions. More specifically, we consider a context in which the word is used and pass it through the BERT module. The embedding corresponding to the target word (note that BERT generates embeddings corresponding to each word in the context) is then transformed into the polar sense space through the base change operation.
Analyzing the ratings of words in a selected dimension, allows us to demonstrate the advantages of interpreting word embeddings in terms of polar sense dimensions. We first consider the dimension ``Black.a.02''$\leftrightarrow$``White.a.02'' (in the sense of \emph{ethnicity}) and transform the embeddings of celebrities and nationalities on this dimension. The observations mostly match the ethnicities of the individuals (see figure~\ref{fig:coldhot}(a)). We also consider words such as milk, coal etc. which are not related to ``Black.a.02''$\leftrightarrow$``White.a.02''  in the sense of \emph{ethnicity} and observe their corresponding scores in this dimension to be neutral. However, their representation on the dimension ``Black.a.01''$\leftrightarrow$``White.a.01'' (in the sense of \emph{color}), captures their semantic well. This demonstrates the benefits of using polar senses as dimensions instead of words, which would have failed to differentiate between the two senses.

We also consider other dimensions and present the connotative meanings of words across these dimensions in figure~\ref{fig:coldhot}(b) which leads to interesting observations. For example, ``politician'', ``meeting'' are more aligned towards ``Hate.a.01'' (in the sense of \emph{disgust}). Similarly, ``murder'' and ``devil'' are aligned towards ``Wrong.a.01'' (in the sense of \emph{morality}).

In addition to picking out interesting dimensions by hand, we also propose to evaluate interpretability by investigating the most descriptive dimensions of a given word. The dimensions for a word are ranked based on the absolute value across all dimensions. Ideally, the top dimensions should be the most descriptive and fitting for the word. For illustration, we provide example words and the corresponding top-5 dimensions in figure~\ref{fig:polar_c_examples}. The top dimensions mostly have a high semantic similarity with the word, and they also reasonably align with human judgement.

\noindent\textbf{Survey experiment:} For evaluating interpretability on a larger scale, we follow the approach by~\citet{mathew2020polar} and conduct a human judgment survey. 
We utilize the crowdsourcing platform Clickworker\footnote{\url{https://www.clickworker.com/}} where we randomly select 15 common English nouns, verbs and adjectives (with short context) and compute their interpretable embedding with \Po.
Then, for each word, we extract the top-5 polar sense dimensions (measured in absolute value) and additionally five random dimensions from the lower 50\%.
These 10 dimensions are then presented to the participants in a random order. Participants are asked to select five dimensions that are most representative of a given word and to rate each dimension based on their alignment to one of the poles on a likert scale between 1 and 7 (with 4 as neutral). Each word is assigned 3 annotators. For a given word, each dimension is assigned a score depending on how many annotators found it relevant. We then select the top 5 dimensions based on this score and we consider them as the ground-truth dimension to which we compare the ones selected by \Po.

In table~\ref{table:conditional}, we present the conditional probability of the top $k$ dimensions selected by \Poo to be also chosen by the human annotators. In the same table, we also report the random chance of getting selected. 
For the top-1 dimension, agreement is roughly 87\% and for the top-2 dimension it is still around 65\%, indicating strong alignment with human judgment.
We also found that the participant's ratings on these dimensions were the (absolute) highest, showing that the word is strongly connected to one of the polar senses.
\begin{table}[h!]
	\centering
	\small
	\begin{tabular}{c|ccccc} 
		
		Top-$k$ & 1 & 2 & 3 & 4 & 5\\
		\hline
		\Poo & 0.876 & 0.558 & 0.312& 0.187& 0.093\\
		\hline
		Random & 0.5 & 0.22 & 0.083& 0.023& 0.004\\
	\end{tabular}
	\caption{Alignment with human judgement. The conditional probability of the top-$k$ dimensions selected by \Poo to be also chosen by the human annotators, together with the random chance of guessing. Significantly higher probabilities than random chance are achieved, indicating that the chosen dimensions are meaningful and match human judgment reasonably well.}
	\vspace{-3mm}
	\label{table:conditional}
\end{table}

\begin{figure}[t]
	\centering
	\includegraphics[scale=0.24]{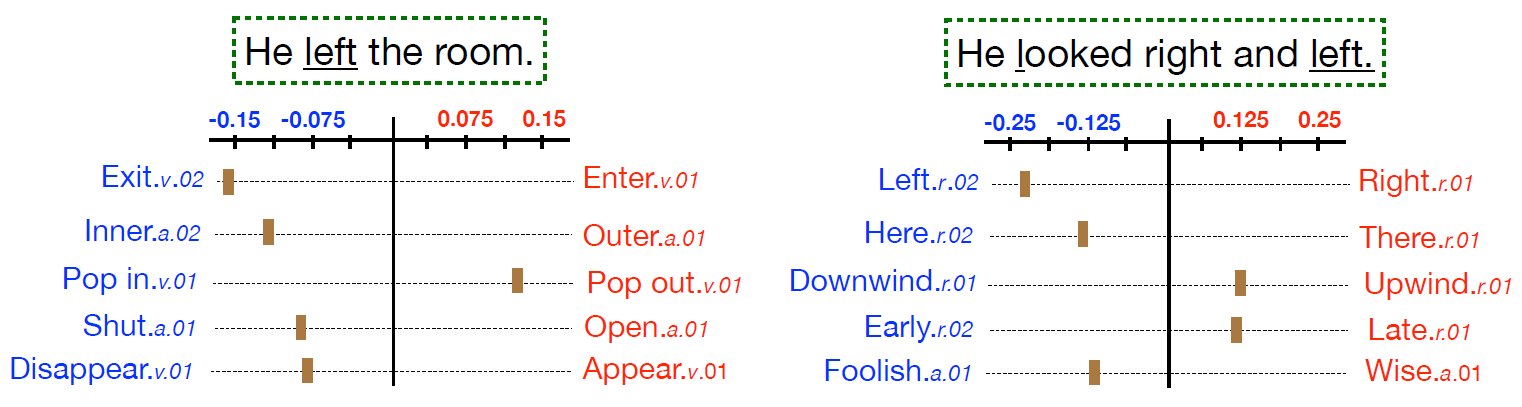}
	\caption{Top-5 dimensions of the word ``left'' for two different contexts in the sense of \emph{going away} (left) and \emph{direction} (right). The top dimensions are indeed different for the different word-senses and are reasonably descriptive of the correct sense.}
	\vspace{-3mm}
	\label{fig:sense}
\end{figure}

\noindent\textbf{Differentiating between senses:} We also evaluate the interpretability of \Poo in terms of its ability to differentiate between two senses of a given word. As an illustrative example, we consider the word ``right'' in the sense of both \emph{direction} and \emph{correctness} (refer to figure~\ref{fig:sense}). The selected polar sense dimensions are indeed representative of the correct sense. Note that the original POLAR framework would not be able to differentiate between the senses, given it generates exactly one embedding for a given word.

We follow up with another human judgement experiment where we present the top-10 polar sense dimensions of words with multiple meanings, together with the word's score on these dimensions, to the annotators.
The task is to identify in which sense the target word is being used in.
We limit this experiment to only two common senses for each word and present the WordNet definitions as the answer possibilities.
Thus, by random guessing, an accuracy of 50\% would be achieved. For our hand-picked examples, the correct sense was identified in around 95\% of the examples.
The average inter-participant agreement on the result is around 78\%. 

The results in this section indicate that \Poo is indeed able to add interpretability to contextual word embeddings and that it aligns reasonably well with human judgement.

\section{Applications}

We discuss two potential use cases: (i) analyzing bias in contextual embeddings and (ii) explaining model decisions on downstream tasks. 

\subsection{Bias Analysis}
Contextual embedding models have been shown to capture bias from training data~\cite{nadeem2020stereoset,ahn2021mitigating,bartl2020unmasking,mozafari2020hate}. We demonstrate that \Poo can be used for analyzing biases by examining the polar sense dimensions which differ the most. 
For our evaluation, we utilize examples from the \textit{Stereoset} dataset from~\citet{nadeem2020stereoset}. 

\noindent\textbf{Next sentence prediction:} BERT can be used to predict the probability of a second sentence following a first one. Consider the sentence pair ``Hispanic people are very diverse. All people like that are criminals.''. Replacing the word ``Hispanic'' by ``American'' leads to significant decrease in BERT's confidence score. We pick the \Poo embeddings of the words ``Hispanic'' and ``American'' from these sentences and compute the polar sense dimensions in which they differ the most. Noticeably, the most discriminative dimensions mostly encode senses related to being or not being a legal citizen such as ``Documented.a.01''$\leftrightarrow$``Undocumented.a.01'' or  ``Irregular.a.01''$\leftrightarrow$``Regular.a.01''. We observe that the word ``Hispanic'' is rated on the side of \textit{irregular}, \textit{undocumented}, \textit{unrighteous} and \textit{unconcerned}, as opposed to ``American''.

\noindent\textbf{Masked LM:} We conduct a similar experiment on the task of masked language modeling. Consider the sentence ``Because he is black, he works as a \verb*|[MASK]|.''. Replacing ``black'' with ``white'' leads to vastly different predictions for the \verb*|[MASK]| token. We apply \Poo to analyze the polar sense dimensions of the \verb*|[MASK]| token in both contexts. We find that the most discriminative dimension is ``Employed.a.01''$\leftrightarrow$``Unemployed.a.01'', indicating that BERT predicts a word more related to \emph{unemployed} when the word ``black'' is used.

\subsection{Explaining classifier results}
\label{sec:expl_cls}
\Poo can further be deployed to explain decisions of classifier models that make use of contextual word embeddings. To illustrate this we consider binary sentiment prediction (positive or negative) on the SST-2 dataset~\cite{socher2013recursive}. We sample and average the \Poo transformed \verb*|[CLS]| tokens from the positive and negative class separately and examine the most discriminative dimensions.  
We find the most discriminative dimensions to be ``sharp''$\leftrightarrow$``dull'', 
``unpleasant''$\leftrightarrow$``pleasant'', 
``endemic''$\leftrightarrow$``cosmopolitan'', ``soft''$\leftrightarrow$``loud'' and ``tasteless''$\leftrightarrow$``tasteful''.
BERT is more likely to classify a review as negative when it is seen as more \textit{sharp}, \textit{unpleasant}, \textit{endemic}, and  \textit{tasteless}.

\section{Discussion }
Next, we discuss issues pertinent to \Po.

\noindent\textbf{Generalizability:} 
\Poo is applicable to any pre-trained contextual embedding model. It can also be deployed on top of any of the constituent transformer layers. This allows for not only comparing different contextual word embedding models in terms of interpretability or bias analysis but also performing similar analysis across transformer layers of the same embedding model. 

\noindent\textbf{Extension to other languages:} 
\Poo should also be extendable to other languages. The only requirement would be to be able to obtain suitable sense antonym pairs as well as example contexts via an oracle. 

\noindent\textbf{Interpretable decision-making.} In section~\ref{sec:expl_cls}, we demonstrated how \Poo could be used to explain decisions of text classifiers. However, the design of \Poo allows for deployment across any other downstream task as well. This is in contrast with existing interpretability methods which are often developed with a particular downstream task in mind. 

\noindent\textbf{Quantitative comparison with other interpretability methods:}  An ideal evaluation set up would have been to quantitatively compare \Poo to other interpretability methods. However, as pointed out in the existing literature~\cite{sundararajan2017axiomatic, sikdar2021integrated}, when two models provide different interpretations, it is difficult to judge if one is better than the other. Involving humans makes it even harder, as one now needs to tease out a person’s own subjective biases. Hence, our crowdsourcing experiments were only designed to understand the efficacy of \Po. Nevertheless, we provide a qualitative comparison with the existing methods in section~\ref{sec:related_work}.

\noindent\textbf{\Poo variants:} Other variants of \Poo can be devised as well. For example, linear transformation instead of base change could be used for obtaining \Poo embeddings. However, we observed that linear transformation does not preserve the original structure of the embedding space, where the different senses of words are already sufficiently separated. 
One can also experiment with different normalization techniques, such as scaling or standardization. In this paper, we concentrated on an exhaustive evaluation setup to include more downstream tasks and crowdsourcing experiments rather than exploring other variants. We consider all the above variants promising avenues for future work.

\section{Related work}
\label{sec:related_work}
In this section, we briefly summarize previous research on enabling interpretability for both static and contextual word embeddings. 

\noindent\textbf{Unsupervised methods:} The key idea for this class of methods is to create sparse embeddings, which is achieved through a post-processing step on top of the  embeddings~\cite{murphy2012learning,faruqui2015sparse,luo2015online}. 
Additionally, the idea of creating sparse embeddings can also be integrated into the word embedding training itself, as demonstrated in~\citet{sun2016sparse,chen2017learning}. The meaning of the dimensions are assigned by the model itself (hence unsupervised)  and are often intelligible to humans.
Notably, Word2Sense~\cite{panigrahi2019word2sense} proposes to create sparse non-negative vectors through Latent Dirichlet Allocation (LDA).
Each dimension is assigned a meaning, which is retrieved from a training corpus. The methods discussed above are specific to static word embeddings. \citet{berend2020sparsity} extends some of these ideas to contextual word embeddings.

\noindent\textbf{(Semi-)supervised methods:} This class of methods aims at adding interpretability to word embeddings by first defining an interpretable space and then transforming the pre-trained embeddings to this space. In this space, each dimension spans between two pole words. While SemAxis \cite{an2018semaxis} proposes to use antonym pairs retrieved from ConceptNet~\cite{speer2017conceptnet}, the POLAR framework~\cite{mathew2020polar} utilizes the semantic differential technique pioneered by Osgood~\cite{osgood1957measurement}. Similarly, BiImp~\cite{csenel2022learning} proposes to use opposite semantic concepts as poles.
Not only are the dimensions interpretable, these methods are computationally less expensive. 


\noindent\textbf{Embedding geometry:} Part of the existing research has focussed on analyzing the position of words in the embedding space. \citet{ethayarajh-2019-contextual} provides evidence that the BERT embeddings are not uniformly distributed in the space, but rather lay on a high dimensional cone. ~\citet{reif2019visualizing} demonstrate that BERT is able to separate fine-grained senses of words by placing them in different locations in space. Similar observations are made by ~\citet{schmidt2020bert} as well. 

\noindent\textbf{Probing:} The goal in probing tasks is to determine whether some syntactic or semantic knowledge is encoded in the produced word embeddings (or attention heads).
The embeddings (or attentions) are fed into a \textit{simple} linear classifier to predict unseen linguistic properties. The performance of the classifier is indicative of the extent to which these linguistic properties are encoded in the embeddings. For BERT, these probing experiments have demonstrated that the layers on the top are more contextual~\cite{ethayarajh-2019-contextual} and the layers at the center contain a large amount of syntactic information~\cite{hewitt-manning-2019-structural,goldberg2019assessing,jawahar2019does,chi2020finding}. The semantic information is generally spread across the entire network~\cite{tenney2019bert,zhao2020quantifying,lin-etal-2019-open}.

\noindent\textbf{Visual explanations:} Finally, recent work has also considered visualizing attention in transformer layers to explain contextual language models~\cite{hoover2020exbert, vig2019multiscale}. Similarly, visualizing word embeddings can also aid in explaining what a model learns as demonstrated in \citet{liu2017visual,heimerl2018interactive,boggust2022embedding} (static) and \citet{sevastjanova2021explaining, berger2020visually} (contextual).

\noindent \textbf{Comparison with \Po:} To the best of our knowledge, \Poo is the first \mbox{(semi-)} supervised method for enabling interpretability for contextual word embeddings. We extend the idea of rating the meaning of words on a scale - defined between two polar \emph{words} - to two polar \textit{senses}. \Poo can also be integrated into task specific fine-tuned models as well. In comparison to \emph{unsupervised} methods, our method enables us to understand the individual dimensions and actively choose and adjust polar sense dimensions for the task at hand. While \emph{probing} and \emph{visualization} methods can reveal whether specific linguistic information is encoded in the embeddings, analyzing the \emph{embedding geometry} can help in uncovering the model characteristics. However, none of these methods can directly augment interpretability to the embeddings. 
Since with \Poo interpretability is directly incorporated into the embeddings, it is applicable to any downstream task. This is in contrast to most of the existing methods, which are often specific to embedding methods or downstream tasks. 

\section{Conclusion}
We introduced \Poo which enables word sense aware interpretability for contextual word embeddings. The key idea is to project word embeddings onto an interpretable space which is constructed from polar sense pairs obtained from an oracle. \Poo extends the original POLAR framework developed for static word embeddings to contextual word embeddings. 
We demonstrated that the obtained interpretable embeddings align well with human judgement. Moreover, \Poo could be integrated into fine-tuned models and can be deployed to specific applications like bias analysis and explaining prediction results of classifier models.

\section{Limitations}
\noindent\textbf{Underlying embedding models:} \Poo uses embeddings of polar senses to build an interpretable subspace. Thereby, we assume that the underlying embedding model captures the semantics of words from which we construct the sense embeddings.
As a result, \Poo is dependent on the quality of the underlying contextual word embedding model. Compared to the original POLAR framework proposed in~\cite{mathew2020polar}, the present approach also depends on the ability of the model to capture individual word-senses with sufficient accuracy.

\noindent\textbf{Presence of bias:} Naturally, our model inherits the biases of the underlying embedding model.
The word ``physics'', for example, has a high rating towards ``male'' on the polar sense scale of ``male'' $\leftrightarrow$ ``female''.
However, \Poo could be used to make these biases visible and potentially help to remove them. One can also tap into state-of-the-art bias mitigation methods (e.g.~\citet{ahn2021mitigating,bartl2020unmasking,mozafari2020hate}) to address this issue.

\noindent\textbf{Dependence on oracles:} The construction of the polar sense space depends largely on the choice of polar opposite senses and the quality of the context examples. Using the example of WordNet, we have shown how a general model can be created. However, we observed that rare senses and low-quality example sentences can lead to poor results.
Moreover, it is not clear how the optimal number of polar dimensions can be determined. Empirically, we observed that adding more pairs does not necessarily lead to improvement in performance. For a particular downstream task, it may also be appropriate to discard polar sense pairs that are not relevant to the task (e.g. if they never occur in the corpus).


\noindent\textbf{Counter-intuitive rating of words.}  We find that in some cases the rating of words on the polar sense scales does not coincide with human judgement. The word ``doctor'', for example, is highly skewed towards ``guilty'' on a scale from ``innocent'' $\leftrightarrow$ ``guilty'', which does not match the typical perception of doctors. We believe this is because word embeddings by design are shaped by their context. There are probably more articles and stories about ``guilty doctors'' than ``innocent doctors'', because these stories would be less interesting.
\section*{Acknowledgements}

Sandipan Sikdar was supported in part by
RWTH Aachen Startup Grant No. \texttt{StUpPD384-20}.


\appendix
\newpage
\section{Appendix}
\label{sec:appendix}
\subsection{Post-processing}
We point out some issues when using WordNet directly as our oracle and present ways to address them.

\subsubsection{Sense Post-processing}\label{sec:problems}
We notice that word-senses identified by WordNet can be overly granular (e.g. according to WordNet there are five different polar sense pairs for ``wet'' $\leftrightarrow$ ``dry'').
To get rid of redundant senses, we propose to use dimension-reduction methods such as \textit{variance maximization} and \textit{orthogonality maximization} from the original POLAR framework~\cite{mathew2020polar}.
Alternatively, one could merge similar senses if the cosine similarity between their sense embeddings is high. Selecting or discarding a rare sense is often task specific.

\subsubsection{Low-quality Example Sentences}
The polar sense embeddings are dependent on the quality of the context (i.e., example sentences demonstrating a particular sense). While deploying WordNet as source for contexts, we encountered some issues which we elaborate on next. 


\noindent\textbf{Flections.} 
When constructing polar sense dimensions, we use the respective word in its basic form and extract the embedding from the example context. However, in WordNet's example sentences, words often do not appear in their basic form, but in inflections (e.g. ``She walk\textbf{s} with a slight limp'' for ``Walk.v.01'').

\noindent\textbf{Synonyms.} 
Occasionally, the word itself is not present in the context sentence but is replaced by a synonym (e.g. ``the \textbf{right} answer'' for ``Correct.a.01'').

\noindent\textbf{Misspellings.} 
We observe that the example sentences often contain spelling mistakes (e.g. ``to\textbf{u}ngued lightning'' for ``Tongued.a.01'')

\noindent\textbf{Mixed-up examples.}
In some cases, the example sentences of a sense are identical to those of the opposite pole (e.g. ``we \textbf{docked} at noon'' for ``Undock.v.01'').

These problems need to be addressed with manual checks of the obtained polar sense pairs and context sentences.

\subsection{Sense Scales}\label{sec:sense_scales}
Instead of rating words on scales defined between two pole words (e.g. ``left'' $\leftrightarrow$ ``right''), our scales are defined between two pole word-senses (e.g. ``left'' $\leftrightarrow$ ``right'' in the sense of \textit{direction}).
While the static POLAR framework rates the word ``correct'' highly on the dimension ``left''  $\leftrightarrow$ ``right'', we expect our framework to rate it low on the dimension ``left''  $\leftrightarrow$ ``right'' in the sense of \textit{direction} but rate it high on the dimension ``wrong'' $\leftrightarrow$ ``right'' in the sense of \textit{correctness}.

To this aim, we analyze whether our constructed representative sense embeddings encode enough sense-related information.
As an illustrative example, we consider the word ``right'' which is used in the senses of \textit{direction}, \textit{correctness} and \textit{lawfulness}.
For each context, we compute the \Poo embeddings for the word and rank the dimensions based on the absolute value. 

\begin{table}[h!]
	\begin{adjustbox}{width=\columnwidth,center}
	\centering
	\begin{tabular}{c|ccc} 
		\hline
		\backslashbox{Sense-Scales}{Context}&he went to the \textbf{right}&his argument is \textbf{right}&film \textbf{rights}\\
		\hline
		Direction\\``left'' $\leftrightarrow$ ``right''&	1$^{st}$&38$^{th}$&32$^{nd}$\\
		\hline
		Correct\\
		``wrong'' $\leftrightarrow$ ``right''&	44$^{th}$&1$^{st}$&291$^{st}$\\
		\hline
		Lawful\\
		``wrong'' $\leftrightarrow$ ``right''&		55$^{th}$&27$^{th}$&9$^{th}$\\
		\hline
	\end{tabular}
	\end{adjustbox}
	\caption{Ability of \Poo in differentiating between senses. We consider the word ``right'' in three different contexts and obtain a ranking of the dimensions for each case. We report the rank of a \Poo dimension in each of the three contexts in each row. For example, the dimension ``left'' $\leftrightarrow$ ``right'' representing the sense of \emph{direction} is ranked first for the context ``he went to the right'', while it is ranked 38th and 32nd respectively in the other two contexts. \Poo is indeed able to identify the correct sense dimensions depending on the context.}
	\label{table:sense_right}
\end{table}
In table~\ref{table:sense_right}, we report the ranks of the polar sense dimensions for each context. 
For the word ``right'' in the context of \textit{direction} ``he went to the right'', the dimension ``left'' $\leftrightarrow$ ``right'' is selected as the most representative dimension (rank 1), while the \textit{correctness} and the \textit{lawful} dimensions are ranked much lower (44 and 55 respectively). Similar results are obtained for the other contexts (see table~\ref{table:sense_right}).

These results indicate that the sense-dimensions of \Poo precisely captures the individual semantics of the senses.

\subsection{Computational Requirements}
\Poo embeddings of words for a given context can be obtained at low cost if the polar sense space is pre-computed. We provide such an implementation along with the submission and encourage readers to review it. Our implementation can even be run on a personal computer. 

Given $n$ polar sense dimensions, inversion of the matrix can be computed in the worst case in $O(n^3)$. Since in our case $n=1762$, the computation is very fast. Moreover, this computation needs to be performed only once.

Retraining the task-specific feed-forward layer with \Poo embeddings was performed on a computing server with 1 TB RAM, 72 cores, each Intel Xeon Gold 6140 CPU at 2.30 GHZ and 2 Tesla P100-PCIE, 16GB GPUs. We would like to reiterate that the retraining is also quite cheap given only the task-specific feed-forward layer needs to be trained. 

\end{document}